\documentclass[10pt,twocolumn,letterpaper]{article}

\usepackage[pagenumbers]{cvpr} 

\usepackage[dvipsnames]{xcolor}

\newcommand{\figref}[1]{Figure~\ref{#1}}

\newcommand{\img}[1]{\mathbf{I}_{#1}}
\newcommand{\depth}[1]{\mathbf{D}_{#1}}
\newcommand{\gauss}[1]{\mathcal{P}_{#1}}
\newcommand{\camext}[1]{\mathbf{P}_{#1}}
\newcommand{\mask}[1]{\mathbf{M}_{#1}}

\newcommand{\normone}[1]{\left\lVert #1 \right\rVert_1}

\newcommand{\paren}[1]{\left( #1 \right)}
\newcommand{\bparen}[1]{\left[ #1 \right]}

\DeclareMathOperator*{\argmin}{argmin}

\usepackage{times}
\usepackage{epsfig}
\usepackage{graphicx}
\usepackage{amsmath}
\usepackage{amssymb}
\usepackage{nicefrac}
\usepackage{pifont}
\usepackage{dsfont}
\usepackage{tabularx}
\usepackage{tabulary}
\usepackage{booktabs}
\usepackage{multirow}
\usepackage{subcaption}
\usepackage{animate}
\usepackage[linesnumbered,ruled,lined,commentsnumbered]{algorithm2e}

\usepackage{color}
\newcommand\blfootnote[1]{%
  \begingroup
  \renewcommand\thefootnote{}\footnote{#1}%
  \addtocounter{footnote}{-1}%
  \endgroup
}

\definecolor{cvprblue}{rgb}{0.21,0.49,0.74}
\usepackage[pagebackref,breaklinks,colorlinks,citecolor=cvprblue]{hyperref}

\def\paperID{}
\def\confName{CVPR}
\def\confYear{2024}

\title{LucidDreamer: Domain-free Generation of 3D Gaussian Splatting Scenes}

\author{Jaeyoung Chung\textsuperscript{*} \qquad Suyoung Lee\textsuperscript{*} \qquad Hyeongjin Nam \qquad Jaerin Lee \qquad Kyoung Mu Lee \\
ASRI, Department of ECE, Seoul National University, Seoul, Korea\\
{\tt\small \{robot0321, esw0116, namhjsnu28, ironjr, kyoungmu\}@snu.ac.kr}
}

\begin{document}
\include{macro}

\twocolumn[{%
\renewcommand\twocolumn[1][]{#1}%
\maketitle
\thispagestyle{empty}
\begin{center}
    \includegraphics[width=0.93\textwidth]{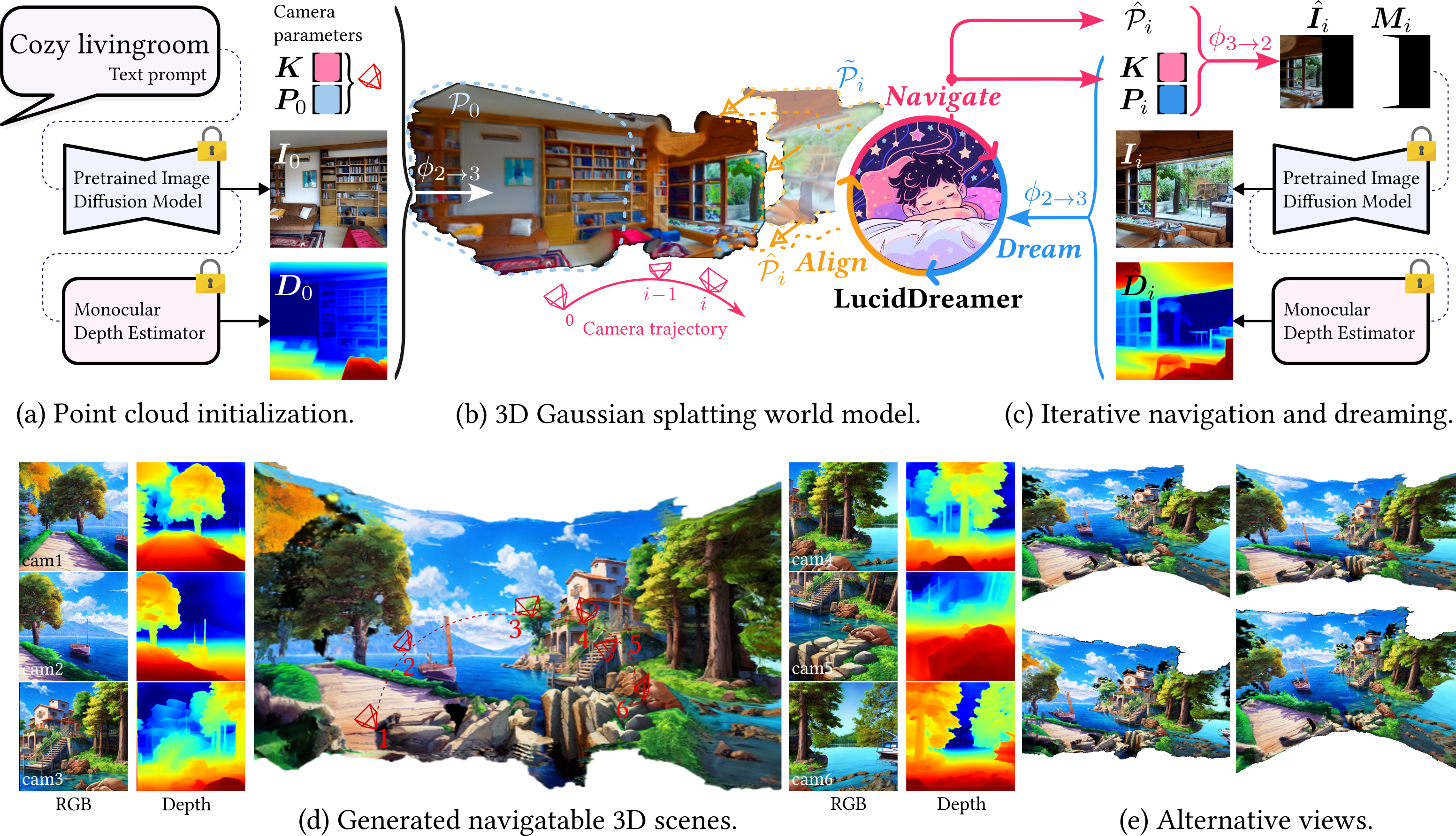}
    \vspace{-0.7mm}
    \captionof{figure}{\textbf{Introducing LucidDreamer.}
    We develop \emph{LucidDreamer}, a general framework for generating multiview-consistent and high-quality 3D scenes from various input types: text, RGB, and RGBD.
    After the initial point cloud is created by lifting the RGBD image, LucidDreamer maintains and expands its world model by repeating two operations: dreaming and alignment.
    The 3D scene is finalized through optimizing a Gaussian splatting representation.
    }
    \label{fig:fig1_teaser}
\end{center}
}]

\blfootnote{\hspace{-0.3cm}\textsuperscript{*} indicates equal contribution.}
\begin{abstract}

With the widespread usage of VR devices and contents, demands for 3D scene generation techniques become more popular.
Existing 3D scene generation models, however, limit the target scene to specific domain, primarily due to their training strategies using 3D scan dataset that is far from the real-world.
To address such limitation, we propose LucidDreamer, a domain-free scene generation pipeline by fully leveraging the power of existing large-scale diffusion-based generative model.
Our LucidDreamer has two alternate steps: Dreaming and Alignment.
First, to generate multi-view consistent images from inputs, we set the point cloud as a geometrical guideline for each image generation.
Specifically, we project a portion of point cloud to the desired view and provide the projection as a guidance for inpainting using the generative model.
The inpainted images are lifted to 3D space with estimated depth maps, composing a new points.
Second, to aggregate the new points into the 3D scene, we propose an aligning algorithm which harmoniously integrates the portions of newly generated 3D scenes.
The finally obtained 3D scene serves as initial points for optimizing Gaussian splats.
LucidDreamer produces Gaussian splats that are highly-detailed compared to the previous 3D scene generation methods, with no constraint on domain of the target scene.
%

\end{abstract}
\vspace{-3mm}
\section{Introduction} \label{sec:intro}

With the advent of commercial mixed reality platforms and the rapid innovations in 3D graphics technology, high-quality 3D scene generation has become one of the most important problem in computer vision.
This requires the ability to create diverse and photo-realistic 3D scenes from any type of input, such as text, RGB, and RGBD images.  
There are efforts to use the diffusion model in voxel, point cloud, and implicit neural representation to generate 3D objects and scenes directly ~\cite{zhou20213d, luo2021diffusion, dupont2022data}, but the results show low diversity and quality due to the limitations in training data based on 3D scans.
One way to cope with the issue is to leverage the power of a pre-trained image generation diffusion model, such as Stable Diffusion~\cite{rombach2022high}, to create diverse high-quality 3D scenes. 
Such a big model creates plausible images with a data-driven knowledge learned from the large-scale training data, although it does not guarantee multi-view consistency between the generated images~\cite{tang2023mvdiffusion}.

In this work, we propose a pipeline called \textbf{LucidDreamer} that utilizes Stable Diffusion~\cite{rombach2022high} and 3D Gaussian splatting\cite{kerbl20233d} to create diverse high-quality 3D scenes from various types of inputs such as text, RGB, and RGBD.
Following the pipeline of LucidDreamer, a unified large point cloud is generated by repeating the two processes named Dreaming and Alignment, alternatively.
Before beginning the two process, an initial point cloud is generated by the initial image and the corresponding depth map.
Dreaming process includes the generation of geometrically consistent images and the lifting of these images into 3D space.
We first move the camera along the pre-defined camera trajectory and project a visible region of point cloud in the new camera coordinate to the new camera plane.
Then, the projected image is put into the Stable Diffusion-based inpainting network to generate the complete image from the projected one.
A new set of 3D points are generated by lifting the inpainted image and the estimated depth map to the 3D space.
Then the proposed alignment algorithm seamlessly connects the new 3D points to the existing point cloud by slightly moving the position of the new points in the 3D space.
After the large point cloud generated by repeating the above processes a sufficient number of time is obtained, we use it as the initial SfM points to optimize the Gaussian splats.
The continuous representation of 3D Gaussian splats removes the holes generated by the depth discrepancy in the point cloud, enabling us to render more photo-realistic 3D scenes than traditional representations.  
\figref{fig:fig1_teaser} shows the simple process of LucidDreamer and a 3D generation result.

LucidDreamer exhibits significantly more realistic and astonishing results compared to existing models.
We compare the generated 3D scenes conditioned with an image from ScanNet~\cite{dai2017scannet}, NYUDepth~\cite{silberman2012indoor}, and Stable Diffusion, and show better visual results across all datasets.
Our model is capable of generating 3D scenes across diverse domains such as realistic/anime/lego and indoor/outdoor. 
Not only does our model support various domains, but it also accommodates the simultaneous use of diverse input conditions.
For example, by conditioning an image and text together, it generates a 3D scene based on the text but also includes the image. 
This alleviates the challenges associated with creating the desired scene solely from the text, moving away from generating samples exhaustively.
Furthermore, our approach also allows for the change of the input condition while creating the 3D space. 
These capabilities offer opportunities to create a wide range of 3D scenes, inspiring creativity.

In summary, our contributions are as follows.
\begin{description}
\item[$\bullet$] We introduce LucidDreamer, a domain-free high-quality 3D scene generation, achieving better domain generalization in 3D scene generation by leveraging the power of Stable Diffusion, depth estimation, and explicit 3D representation.
\item[$\bullet$] To generate multi-view images from Stable Diffusion, our \textbf{Dreaming} process establishes point cloud as geometrical guideline for each image generation. 
Subsequently, our \textbf{Aligning} process harmoniously integrates the generated images to form an unified 3D scene.
\item[$\bullet$] Our model provides users with the ability to create 3D scenes in various ways by supporting different input types, such as text, RGB, and RGBD, allowing the simultaneous use of multiple inputs, and enabling the change of the inputs during the generation process.
\end{description}
\section{Related Work} \label{sec:relatedwork}

\paragraph{3D Scene Representation.}
Representative methods for expressing 3D scenes include explicit methods such as point cloud, mesh, and voxel.
These are widely used because they allow direct and intuitive control of each element and enable fast rendering through the rasterization pipeline.
However, they need a large number of elements for detailed expression because of their simple structure. Complex primitives such as cuboid~\cite{tulsiani2017learning}, Gaussian~\cite{foygel2010extended}, ellipsoid~\cite{genova2019learning}, superquadrics~\cite{paschalidou2019superquadrics}, convex hull~\cite{deng2020cvxnet}, and polynomial surface~\cite{yavartanoo20213dias} were developed for more efficient expression.
Although primitives have increased expressive power for complex geometry, it is still difficult to express realistic 3D scenes because of simple color representation.

Recently, there have been works to express more detailed 3D scenes using neural networks as implicit expressions.
They train a neural network to express the scene creating the desired properties in 3D coordinates, such as signed distance function~\cite{park2019deepsdf, takikawa2021neural}, RGB$\alpha$~\cite{mildenhall2021nerf, sitzmann2020implicit}.
In particular, Neural Radiance Fields~\cite{mildenhall2021nerf} showed that it was possible to optimize photorealistic 3D scenes from multiple images through volume rendering, but the scene implicitly stored in the network form is difficult to handle and slow.
To improve this, subsequent studies attempted to use volume rendering in explicit expressions. 
By utilizing the locality of structures such as sparse voxels\cite{liu2020neural, yu2021plenoctrees, sun2022direct, fridovich2022plenoxels}, featured point clouds\cite{xu2022point}, Multi-Level Hierarchies~\cite{muller2021real, muller2022instant}, tensor~\cite{chen2022tensorf}, infinitesimal networks~\cite{garbin2021fastnerf, reiser2021kilonerf}, triplane~\cite{chan2022efficient}, polygon~\cite{chen2023mobilenerf}, and Gaussain splats~\cite{kerbl20233d} they greatly improve the training and rendering speed. 
In particular, 3D Gaussian splatting~\cite{kerbl20233d} utilizes the concept of Gaussian splats combined with spherical harmonics and opacity to represent complete and unbounded 3D scenes.
It supports not only alpha-blending but also differentiable rasterization, resulting in fast, high-quality 3D scene optimization.
This structure is essential for our generation method, which cannot determine the bounds of the scene due to sequential image generation, and plays a role in making the scene complete.

\paragraph{3D Scene Generation.}
Inspired by the early success of generative adversarial network (GAN)~\cite{goodfellow2014generative} in image generation, similar attempts are made in 3D creation. 
Creating a set of multiview consistent images~\cite{chan2021pi, schwarz2020graf, niemeyer2021giraffe}, or directly creating voxel~\cite{wu2016learning, nguyen2019hologan, nguyen2020blockgan} or point cloud~\cite{shu20193d, achlioptas2018learning} were studied.
However, they suffer from GAN's learning instability~\cite{radford2015unsupervised} and memory limitation in 3D representation, limiting the generation quality.
Encouraged by the recent success of diffusion~\cite{ho2020denoising, song2020denoising} in the field of image generation~\cite{rombach2022high, ramesh2021zero}, there are many attempts to introduce the diffusion model into 3D representation, such as voxel~\cite{zhou20213d}, point cloud~\cite{luo2021diffusion, zeng2022lion}, triplane~\cite{chan2022efficient, shue20233d, chen2023single}, implicit neural network~\cite{dupont2022data, you2023generative, poole2022dreamfusion}.
They use object-centric coordinates because of their nature and focus on simple examples.
Some generative diffusion models overcome this problem by using a mesh as a proxy and diffusing in the UV space. 
They create a large portrait scene by continuously building the mesh~\cite{fridman2023scenescape} or create indoor scenes~\cite{lei2023rgbd2, cohen2023set} and more realistic objects~\cite{qian2023magic123, tang2023dreamgaussian}.
However, their performance falls short of foundation models\cite{rombach2022high} because they involve training a new diffusion model in a different representation space, which is limited by data availability and computational resources.
In comparison, our method leverages the power of the foundation model to generate diverse images and creates reliable 3D scenes through depth estimation and optimization.

\section{Method} \label{sec:method}

While the range of target scenes of existing scene generation models is strictly restricted due to the limitations in training dataset, LucidDreamer can generate even more realistic, higher-resolution 3D scenes with much more general input conditions.
For instance, LucidDreamer can generate a text-relevant scene if only the text prompt is given.
Also, the style of the input image is maintained along the scene, while existing models keep producing scenes that are similar to the style of the training dataset, not the input image.

The pipeline of LucidDreamer is broadly divided into two stages: point cloud construction and Gaussian splats optimization.
During the first stage, an initial point cloud is formed from the input image, and the area of the point cloud is expanded to create a large scene using Stable Diffusion inpainting and monocular depth estimation.
Then, the point cloud and the reprojected images are used to optimize Gaussian splats.
By representing the scene with Gaussian splats, we can fill the empty space that appears in the point cloud due to the depth discrepancy.

\subsection{Point cloud construction}
\label{sec:constgauss}

To generate multi-view consistent 3D point cloud, we create the initial point cloud and aggregate the points by moving back and forth between 3D space and the camera plane while moving the camera.
The overall process of point cloud construction is illustrated in \figref{fig:fig1_teaser}.

\paragraph{Initialization.}
A point cloud generation starts from lifting the pixels of the initial image.
If the user gives a text prompt as input, the latent diffusion model is used to generate an image relevant to the given text, and the depth map is estimated using the monocular depth estimation model such as ZoeDepth~\cite{bhat2023zoedepth}.
We denote the generated or received RGB image and the depth map as $\img{0} \in \mathbb{R}^{3 \times H \times W}$ and $\depth{0} \in \mathbb{R}^{H \times W}$, where $H$ and $W$ are height and the width of the image.
The camera intrinsic matrix and the extrinsic matrix of $\img{0}$ are denoted as $\mathbf{K}$ and $\camext{0}$, respectively.
For the case where $\img{0}$ and $\depth{0}$ are generated from the diffusion model, we set the values of $\mathbf{K}$ and $\camext{0}$ by convention regarding the size of the image.

From the input RGBD image $[\img{0}, \depth{0}]$, we lift the pixels into the 3D space, where the lifted pixels will form a point cloud in a 3D space.
The generated initial point cloud using the first image is defined as $\gauss{0}$:

\begin{equation}
    \gauss{0} = \phi_{2 \rightarrow 3} \left( [\img{0}, \depth{0}], \mathbf{K}, \camext{0} \right),
    \label{eq:firstgauss}
\end{equation}
where $\phi_{2 \rightarrow 3}$ is the function to lift pixels from the RGBD image $[\mathbf{I}, \mathbf{D}]$ to the point cloud.

\vspace{-2mm}
\paragraph{Point cloud aggregation.}
We sequentially attach points to the original point cloud to create a large 3D scene.
Specifically, we set the camera trajectory with length $N$, where $\camext{i}$ indicates the position and pose of the camera in the $i$-th index, then inpaint and lift the missing pixel in each step.
Here, the generated points should satisfy two conditions; the images projected from the points should have high perceptual quality and be consistent with image parts produced from the existing points.
To achieve the former condition, we borrow the representation power of the Stable Diffusion~\cite{rombach2022high} to the image inpainting task.

\textbf{Navigation}. At step {i}, we first move and rotate the camera from the previous position ($\camext{i-1}$) to $\camext{i}$.
We change the coordinate from the world to the current camera and project to the camera plane using $\mathbf{K}$ and $\camext{i}$.

\textbf{Dreaming}. We denote the projected image at camera $\camext{i}$ as $\hat{\mathbf{I}}_{i}$.
Since the position and the pose of the camera are changed, there would be some regions that cannot be filled from the existing point cloud.
We define the mask $\mask{i}$ to discriminate the region that is filled by existing points in $\hat{\mathbf{I}}_{i}$.
Specifically, the value of $\mask{i}$ is one if the corresponding pixel is already filled or 0 otherwise.
The Stable Diffusion inpainting model ($\mathcal{S}$) is executed to generate a realistic image, $\img{i}$, from the incomplete image ($\hat{\mathbf{I}}_{i}$) and the mask ($\mask{i}$).
The corresponding depth map ($\hat{\mathbf{D}}_{i}$) is estimated using the monocular depth estimation network ($\mathcal{D}$).
Here, the monocular depth estimation model can only estimate the relative depth, and the depth coefficients from the relative depth to the actual depth can be different between images.
If the depth coefficients are different, the lifted 3D point clouds in the two generated images are not connected and are spaced apart.
We estimate the optimal depth scale coefficient, $d_i$, that minimizes the distance between the 3D points of the new image and the corresponding points in the original point cloud, $\gauss{i-1}$.
Then the actual depth map, $\depth{i}$ is calculated by multiplying the coefficient $d_i$ to the estimated depth map, $\hat{\mathbf{D}}_{i}$.
\begin{equation}
\begin{split}
    \img{i} = \mathcal{S} & \paren{ \hat{\mathbf{I}}_{i}, \mask{i} }, 
    \hat{\mathbf{D}}_{i} = \mathcal{D} \paren{ \img{i} },
    \depth{i} = d_i \hat{\mathbf{D}}_{i}, \\
    d_i = \argmin_{d} & \left( \sum_{ \mask{i} = 1} \normone{\phi_{2 \rightarrow 3} \paren{ \bparen{\img{i}, d \hat{\mathbf{D}}_{i}}, \mathbf{K}, \camext{i}} - \gauss{i-1} } \right).
\end{split}
\label{eq:genimgdep}
\end{equation}
Here, $ \mask{i} = 1$ implies that the distance of point pairs in the overlapping regions is used for estimating $d_i$.

Using the image and the corresponding depth map, $\bparen{\img{i}, \depth{i}}$, we lift the pixels to 3D space.
Here, we note that only inpainted pixels of $\img{i}$ are lifted to prevent points overlapping and mitigate the inconsistency problem.
The output of dreaming, $\hat{\mathcal{P}}_{i}$, can be calculated as:

\begin{equation}
        \hat{\mathcal{P}}_{i} = \phi_{2 \rightarrow 3} \paren{ \bparen{\img{i}, \depth{i}|\mask{i}=0}, \mathbf{K}, \camext{i}},
\label{eq:lift}
\end{equation}
where $\bparen{\img{i}, \depth{i}|\mask{i}=0}$ indicates the inpainted region in the RGBD image.

\begin{algorithm}[t]
    \SetKwData{Left}{left}\SetKwData{This}{this}\SetKwData{Up}{up}
    \SetKwFunction{Union}{Union}\SetKwFunction{FindCompress}{FindCompress}
    \SetKwInOut{Require}{Require}
    \SetAlgoLined
    \caption{Constructing point cloud}
    \label{al:constpcd}
    \KwIn {A single RGBD image $\bparen{ \img{0}, \depth{0} }$}
    \KwIn {Camera intrinsic $\mathbf{K}\,$, extrinsics $\{\mathbf{P}_i\}_{i=0}^{N}$}
    \KwOut {Complete point cloud $\mathcal{P}_N$}
    \BlankLine
    $\gauss{0} \gets \phi_{2 \rightarrow 3} \left( [\img{0}, \depth{0}], \mathbf{K}, \camext{0} \right)$ \\
    \For{$i\gets1$ \KwTo $N$}{
        $\hat{\mathbf{I}}_{i}, \mask{i} \gets \phi_{3 \rightarrow 2}\paren{ \gauss{i-1}, \mathbf{K}, \camext{i}}$ \\
        $\mathbf{I}_{i} \gets \mathcal{S}  \paren{ \hat{\mathbf{I}}_{i}, \mask{i} }\,$, 
        $\hat{\mathbf{D}}_{i} \gets \mathcal{D} \paren{ \mathbf{I}_{i} }$ \\
        $d_i \gets 1$ \\
        \While{not converged}{
            $\tilde{\mathcal{P}}_{i} \gets \phi_{2 \rightarrow 3} \paren{ \bparen{\img{i}, d_i \hat{\mathbf{D}}_{i}}, \mathbf{K}, \camext{i}}$ \\
            $\mathcal{L}_{d} \gets \frac{1}{\lVert \mask{i} = 1 \rVert}\sum_{ \mask{i} = 1} \normone{\tilde{\mathcal{P}}_{i} - \gauss{i-1}}$ \\
            Calculate $\nabla_{d} \mathcal{L}_{d}$ \\
            $d_i \gets d_i - \alpha \nabla_{d} \mathcal{L}_{d}$
        }
        $\depth{i} \gets d_i \hat{\mathbf{D}}_{i}$ \\
        $\hat{\mathcal{P}}_{i} \gets \phi_{2 \rightarrow 3} \paren{ \bparen{\img{i}, \depth{i}|\mask{i}=0}, \mathbf{K}, \camext{i}}$ \\
        $\gauss{i} \gets \gauss{i-1} \cup \mathcal{W}\paren{ \hat{\mathcal{P}_i} }$
    }
\end{algorithm}
\begin{figure*}[t]
    \centering

  \centering
    \begin{minipage}[]{0.03\linewidth}%
            \scalebox{0.9}{\rotatebox{90}{%
            \begin{tabular}[t]{>{\centering\arraybackslash}p{0.5cm}
                               >{\centering\arraybackslash}p{3.6cm}
                               >{\centering\arraybackslash}p{4.2cm}
                               }
                 & (ii) RGB & 
                (i) Text
            \end{tabular}
            }}
        \end{minipage}
    \begin{minipage}[]{0.95\linewidth}
        \centering
        \includegraphics[width=\linewidth]{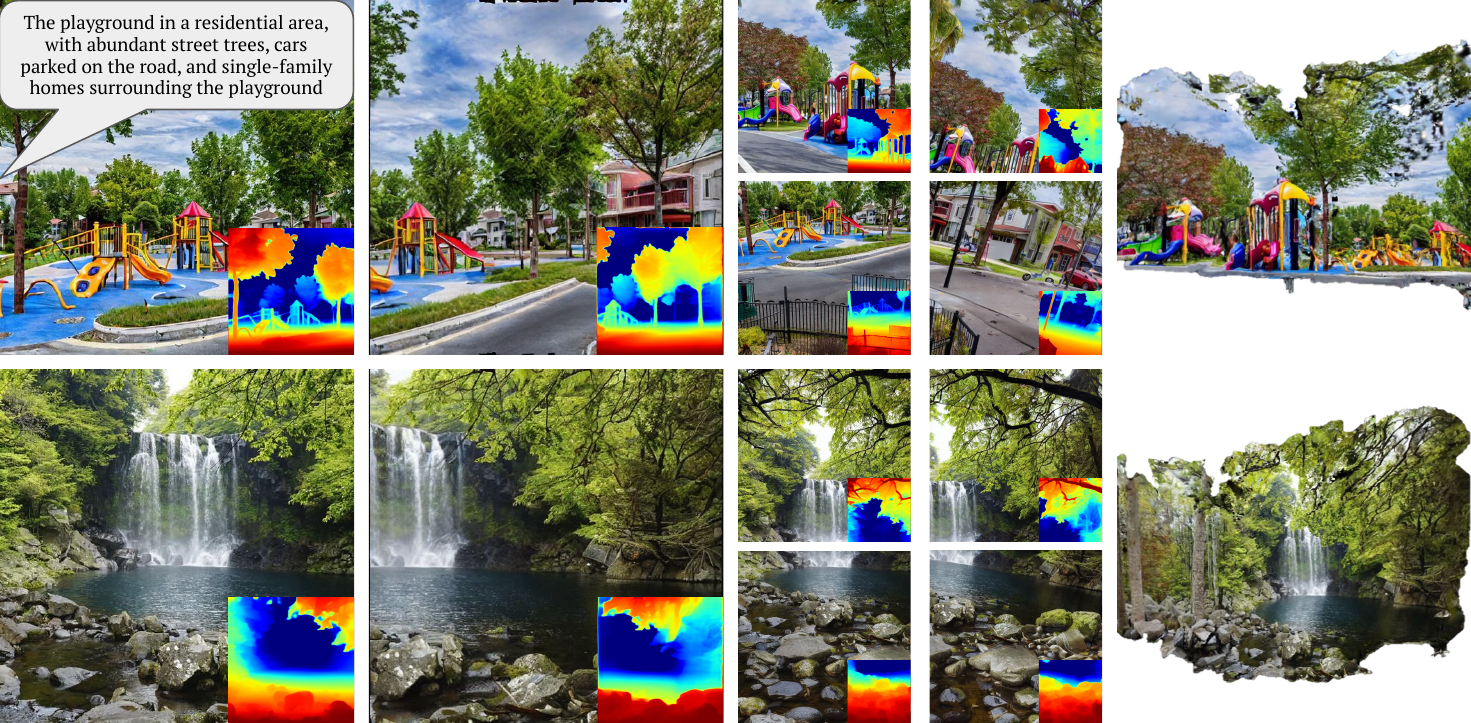}

        \vspace{-3.5mm}
        \subfloat[input($\img{0}$)\label{fig:multi-i0}]{\includegraphics[width=0.24\linewidth]{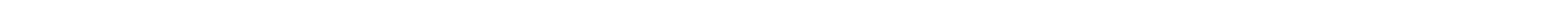}}
        \hfill
        \subfloat[$\img{1}$\label{fig:multi-i1}]{\includegraphics[width=0.24\linewidth]{sec/figure/blank.png}}
        \hfill
        \subfloat[$\img{i}$s\label{fig:multi-is}]{\includegraphics[width=0.25\linewidth]{sec/figure/blank.png}}
        \hfill
        \subfloat[Generated 3D\label{fig:multi-3d}]{\includegraphics[width=0.24\linewidth]{sec/figure/blank.png}}
        \\

    \end{minipage}
    \vspace{-2mm}
    \caption{
        \textbf{Intermediate images during point cloud generation and final 3D output between different inputs.}
        We generate 3D scene from different input types (text and RGB image).
        The input image in the first row is generated image using Stable diffusion.
        Our model is capable of generating consistent images high-quality 3D scene regardless of input type.
    }
    \vspace{-3mm}
    \label{fig:multiinput}
\end{figure*}

\textbf{Alignment}. Compared to the way that trains a generative model to generate both RGB and the depth map at once, such as RGBD2~\cite{you2023generative}, the depth map estimated by off-the-shelf depth estimation method is more accurate and generalizable to various situations since off-the-shelf methods are trained on large and various datasets.
However, since $\depth{0}, \depth{1}, ..., \depth{i-1}$ is not considered when estimating $\depth{i}$, inconsistency problem occurs when we add new points, $\hat{\mathcal{P}}_{i}$.
To overcome the problem, we move the points of $\hat{\mathcal{P}}_{i}$ in 3D space to attach the two point cloud ($\gauss{i-1}$ and $\hat{\mathcal{P}}_{i}$) smoothly.
Specifically, we extract the region where the value of mask changes ($\lvert \nabla \mask{i} \rvert > 0$) to find the corresponding points to that region in both $\gauss{i-1}$ and $\hat{\mathcal{P}}_{i}$.
Then, we calculate the displacement vector from $\hat{\mathcal{P}}_{i}$ to $\gauss{i-1}$.
However, moving the points in a naive way may distort the shape of the lifted point cloud and make a misalignment between the point cloud and the inpainted image.
We mitigate the issue by giving the restrictions for moving the points and using the interpolation algorithm to preserve the overall shape of the points.

First, we force each point in $\hat{\mathcal{P}}_{i}$ to move along the ray line from the camera center to the corresponding pixel.
We find the closest point to the corresponding point in $\gauss{i-1}$ along the ray line and report how much the depth changes are caused by the movement.
Using the constraint, we preserve the contents of RGB image~($\img{i}$) although moving the points in 3D space.
Next, we assume that the depth does not change at the opposite side of the mask boundary region.
Then, for the points that do not have their ground truth counterparts, \textit{i.e.} $\mask{i} = 0$, we calculate for each pixel how much the depth value should change using linear interpolation.
By interpolating smoothly, the mismatch among the pixels caused by the drastic movement is alleviated.
The aligned points are combined with the original one:

\begin{equation}
    \gauss{i} = \gauss{i-1} \cup \mathcal{W}\paren{ \hat{\mathcal{P}_i} }, 
    \label{eq:union}
\end{equation}
where we denote calculating movement and interpolation as $\mathcal{W}$.
We repeat the process N times to construct the final point cloud, $\gauss{N}$.
By reprojection, $\gauss{N}$ provides high-quality and multi-view consistent images.
The whole process of constructing $\gauss{N}$ from $\bparen{\img{0}, \depth{0}}$, $\mathbf{K}$, and $\{ \mathbf{P}_{i} \}_{i=0}^{N}$ is written in Algorithm~\ref{al:constpcd}.

\subsection{Rendering with Gaussian Splatting}
\label{sec:gaussdepth}

After the point cloud is created, we train 3D Gaussian splatting model~\cite{kerbl3Dgaussians} using the point cloud and the projected images.
The center of Gaussian splatting points are initialized by the input point cloud, and the volume and the position of each point are changed by the supervision of input ground truth projected images.
We use the generated point cloud ($\gauss{N}$) as the initial SfM points.
Initialization with $\gauss{N}$ will boost up the convergence of the network and encourage the network to focus on generating the details of the representation.
For the images to train the model, we use additional $M$ images as well as $(N+1)$ images for generating the point cloud, since the initial $(N+1)$ images are not sufficient to train the network for generating the plausible output.
The $M$ new images and the masks are generated by reprojecting from the point cloud $\gauss{N}$ by a new camera sequence of length $M$, denoted as $\camext{N+1}, ..., \camext{N+M}$.

\begin{equation}
    \img{i}, \mask{i} = \phi_{3 \rightarrow 2}\paren{ \gauss{N}, \mathbf{K}, \camext{i}}, i=M+1, ..., M+N.
    \label{eq:reprojection}
\end{equation}
We note that we do not inpaint $\img{i}$ when optimizing Gaussian splats.
Instead, when calculating the loss function, we only consider the valid image region where the mask value is 1.
It prevents the model from learning the wrong details of the reprojected images.
Since each point is represented as a Gaussian distribution, the missing pixels when training the model are naturally filled, and the rasterized image after the training becomes plausible.

\section{Experiments}
\label{sec:experiment}

\begin{figure*}[t]
    \centering
        \centering
        \subfloat[RGB condition\label{fig:difftext-i0}]{\includegraphics[width=0.24\linewidth]{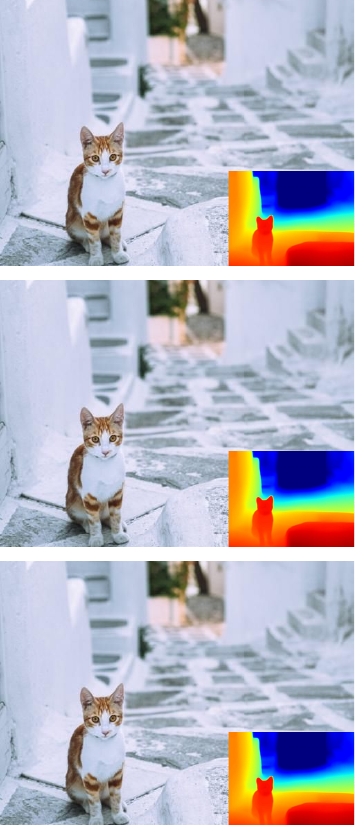}}
        \hfill
        \subfloat[Generation process with different text prompt\label{fig:difftext-text}]{\includegraphics[width=0.5\linewidth]{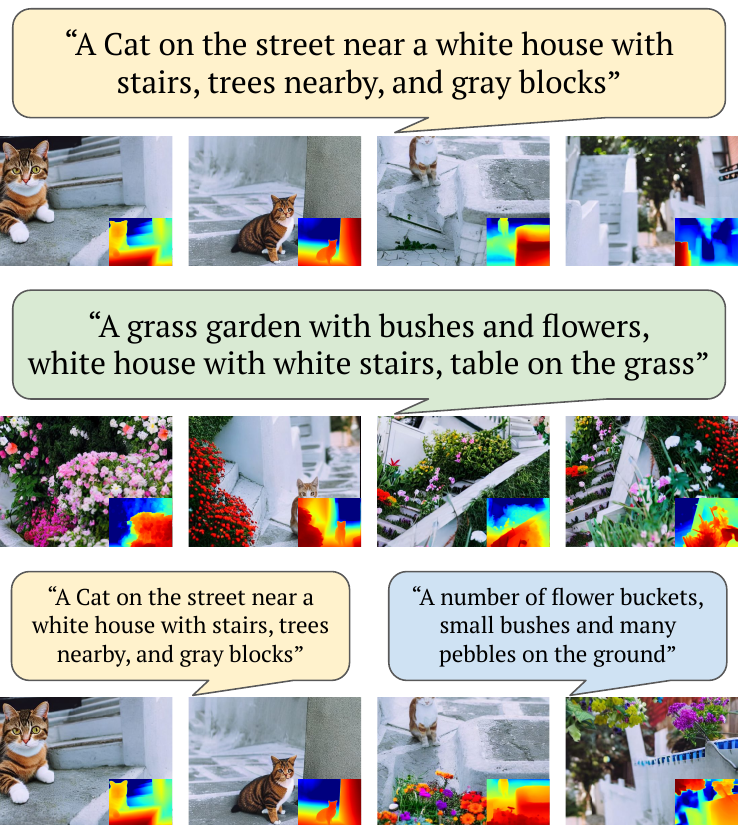}}
        \hfill
        \subfloat[Generated 3D\label{fig:difftext-3d}]{\includegraphics[width=0.24\linewidth]{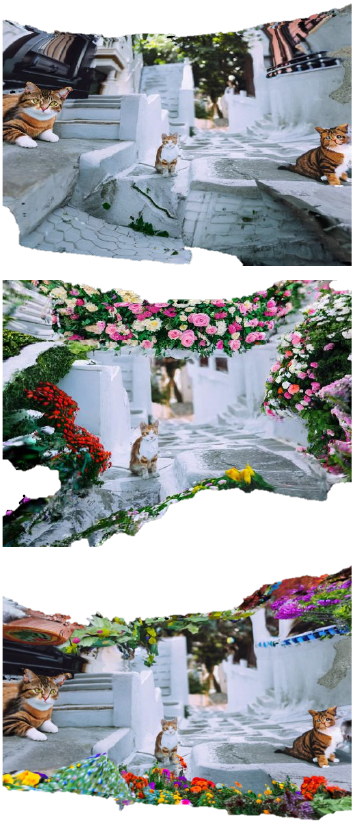}}
    \caption{
        \textbf{Intermediate images during point cloud generation and final 3D output for different text prompt.}
        We put the different text prompt while having same initial image ($\img{0}$) and compare the generation results.
    }
    \vspace{-2mm}
    \label{fig:difftext}
\end{figure*}

\subsection{Experiment settings}

\paragraph{Datasets.}
Since LucidDreamer is optimized for every input, we do not need training dataset to train the model.
For the text input, we randomly generate several text prompts relevant to scene images to generate the first image using Stable Diffusion.
We use real or generated high-quality images for the RGB input.
For the case of RGBD inputs, we use ScanNet~\cite{dai2017scannet} and NYUdepth~\cite{silberman2012indoor} since the two datasets have ground truth depth maps.

\paragraph{Implementation details.}
The modules we used to construct LucidDreamer can be either trained using manual design or brought from off-the-shelf models.
We use pre-trained large-scale off-the-shelf models to compose the whole network to maximize the generalization capability of the network.
Specifically, we adopt Stable Diffusion model~\cite{rombach2022high} to inpaint the masked image.
We use the same text prompt input for the Stable Diffusion if the first image is generated from the text.
If the input format is a RGB(D) image without text, we use LAVIS~\cite{li-etal-2023-lavis} to generate the caption according to the image and place it in the diffusion inpainting model to generate consistent content.
For the camera trajectory that we use to construct the point cloud ($\{\textbf{P}_{i}\}_{i=0}^{N}$), we create several types of camera trajectory presets in advance, and different types of trajectories were used for different tasks.

\subsection{Experiment results}
\begin{figure*}[t]
    \centering

  \centering
    \begin{minipage}[]{0.02\linewidth}%
            \scalebox{0.9}{\rotatebox{90}{%
            \begin{tabular}[t]{>{\centering\arraybackslash}p{0.2cm}
                               >{\centering\arraybackslash}p{6.5cm}
                               >{\centering\arraybackslash}p{6.5cm}
                               >{\centering\arraybackslash}p{6.5cm}
                               }
                 & (iii) NYUDepth~\cite{silberman2012indoor}
                 & (ii) ScanNet~\cite{dai2017scannet}
                 & (i) Generated
            \end{tabular}
            }
        }
        \end{minipage}
    \begin{minipage}[]{0.02\linewidth}%
            \scalebox{0.9}{\rotatebox{90}{%
            \begin{tabular}[t]{>{\centering\arraybackslash}p{0.2cm}
                               >{\centering\arraybackslash}p{3.2cm}
                               >{\centering\arraybackslash}p{3.2cm}
                               >{\centering\arraybackslash}p{3.2cm}
                               >{\centering\arraybackslash}p{3.2cm}
                               >{\centering\arraybackslash}p{3.2cm}
                               >{\centering\arraybackslash}p{3.2cm}
                               }
                 & Ours & RGBD2~\cite{lei2023rgbd2}
                 & Ours & RGBD2~\cite{lei2023rgbd2}
                 & Ours & RGBD2~\cite{lei2023rgbd2}
            \end{tabular}
            }
        }
        \end{minipage}
    \begin{minipage}[]{0.95\linewidth}
        \centering
        \subfloat[input($\img{0}$)\label{fig:rgbd2-i0}]{\includegraphics[width=0.248\linewidth]{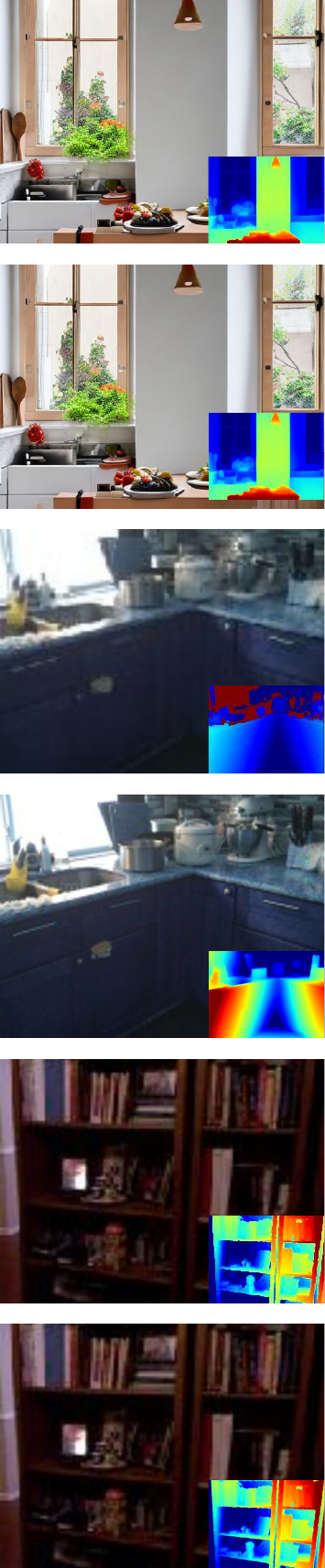}}
        \hfill
        \subfloat[$\img{1}$\label{fig:rgbd2-i1}]{\includegraphics[width=0.248\linewidth]{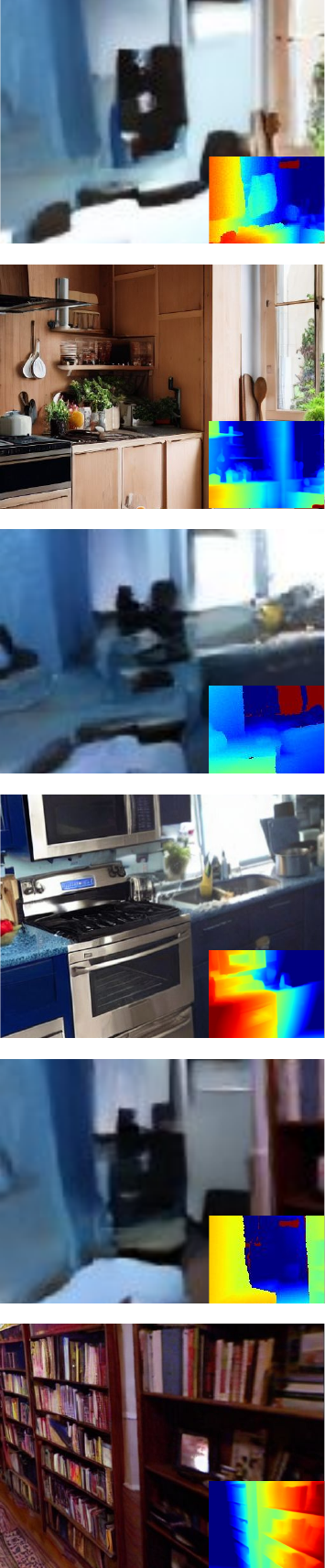}}
        \hfill
        \subfloat[$\img{i}$s\label{fig:rgbd2-is}]{\includegraphics[width=0.248\linewidth]{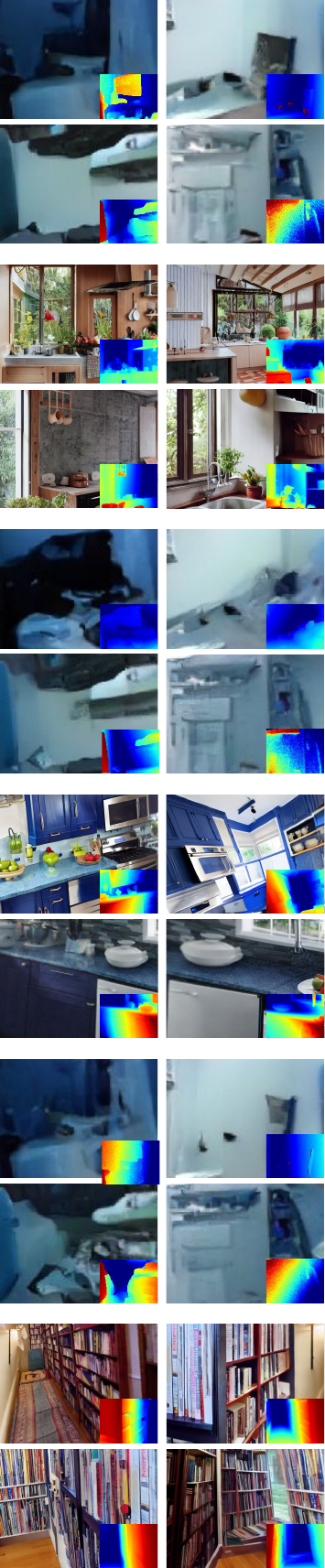}}
        \hfill
        \subfloat[Generated 3D\label{fig:rgbd2-3d}]{\includegraphics[width=0.248\linewidth]{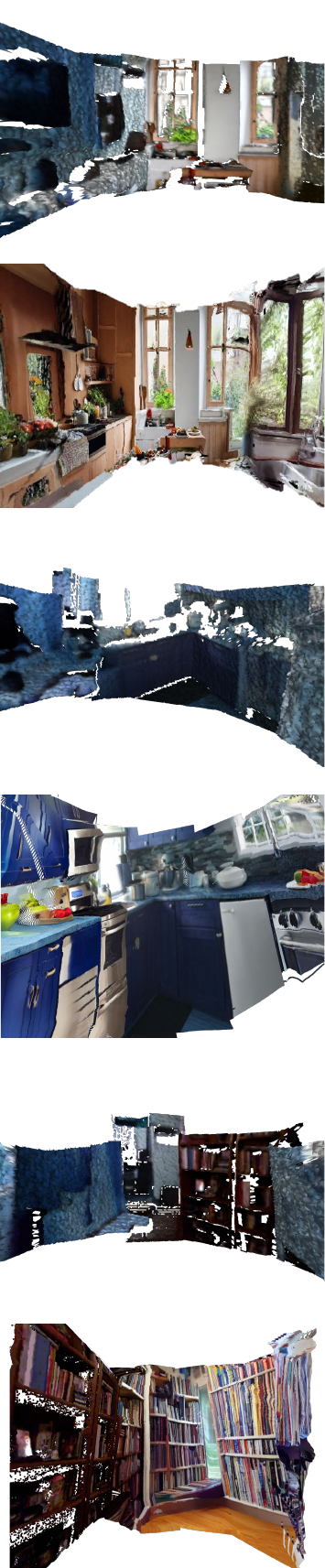}}
        \\

    \end{minipage}
    \caption{
        \textbf{Qualitative comparison with RGBD2~\cite{lei2023rgbd2} on various image datasets.}
        We compare LucidDreamer with RGBD2 starting from the same input image while changing the datasets.
        The scene generated by LucidDreamer always shows higher quality than RGBD2, even on ScanNet~\cite{dai2017scannet} which RGBD2 is trained on.
    }
    \vspace{-1mm}
    \label{fig:comparergbd2}
\end{figure*}

We demonstrate the superiority and high generalizability of LucidDreamer in many aspects.
We strongly recommend the readers to watch the video in the supplementary materials where we can entirely show the strength of our model.

\begin{figure}[t]
    \centering
    \vspace{-2mm}
    \renewcommand{\wp}{0.48 \linewidth}
    \addtocounter{subfigure}{-2}
    \subfloat{
        \centering
        \animategraphics[width=\wp, poster=11, loop, autoplay, final, nomouse, method=widget]{1}{sec/figure/diffdomain_animate1/}{000000}{000022}
    }
    \hfill
    \subfloat{\includegraphics[width=0.48\linewidth]{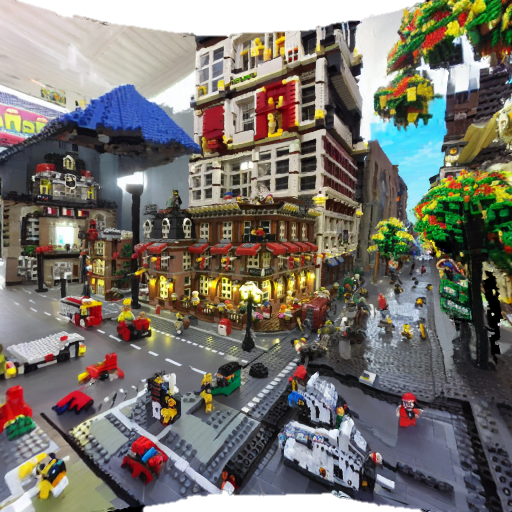}}
    \\
    
    \vspace{1mm}
    \addtocounter{subfigure}{-2}
    \subfloat[Generated video]{
        \centering
        \animategraphics[width=\wp, poster=0, loop, autoplay, final, nomouse, method=widget]{2}{sec/figure/diffdomain_animate2/}{000000}{000010}
    }
    \hfill
    \subfloat[3D whole view]{\includegraphics[width=\wp]{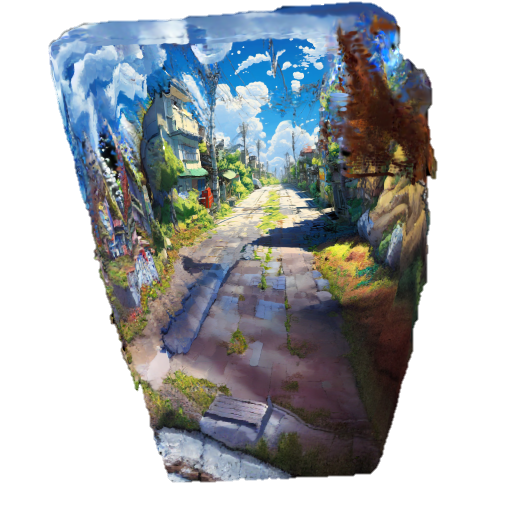}}
    \\
    
    \vspace{1mm}
    \caption{
        \textbf{3D reconstruction results and short video on various styles.} \emph{This is a video figure that is best viewed by Adobe Reader.}
    }
    \vspace{-3mm}
    \label{fig:diffdomain}
\end{figure}

\paragraph{Applicability to various input domains and formats.}
LucidDreamer is capable of generating a consistent and high-quality 3D scene considering the input style.
\figref{fig:multiinput} shows the generated realistic images and the 3D scenes.
At the top row, we visualize a result of Text-to-3D. 
We depict an initial image generated from the given text and estimated depth in (a). 
(b) and (c) present the plausible images and geometry generated through our pipeline involving navigation, dreaming, and alignment.
We showcase an overview of the final 3D scene in (d).
On the other hand, the bottom row demonstrates an example result of RGB-to-3D
We estimated depth from the given RGB and used it as an initial geometry for the scene. 
Similar to the top row, we generated believable images and geometry, resulting in a high-quality 3D scene.
Since our model supports multiple inputs, it can generate 3D scenes in various ways as illustrated in \figref{fig:difftext}.
The top and middle rows depict outcomes generated by guaranteeing the inclusion of conditioned RGB during the creation of the 3D scene.
Despite different texts, the conditioned RGB is consistently present in the scene. 
On the other hand, the bottom row displays the outcome of altering the text condition while generating the 3D scene.
Through diverse combinations and alterations of conditions, our model facilitates the creation of the desired 3D scene more effortlessly.
We illustrate additional example scenes in \figref{fig:diffdomain}. 
Our model successfully generates diverse 3D scenes with various styles (e.g. lego, anime) across different camera paths.

\paragraph{Comparison with RGBD2.}
We qualitatively compare the generation results with RGBD2~\cite{lei2023rgbd2} and illustrate the result in \figref{fig:comparergbd2}.
For fairness of comparison, we compare the results on three images with different domains: generated image, ScanNet, and NYUDepth.
For the generated image, the depth map estimated by Zoedepth~\cite{bhat2023zoedepth} is considered a ground-truth depth map when processing RGBD2.
For ScanNet and NYUDepth, we use the ground truth depth map for both RGBD2 and LucidDreamer when producing a 3D scene.
For ScanNet, each scene consists of several images and the corresponding depth maps and camera views.
We randomly select one of the given image and depth map pairs and use it as an initial RGBD input.
\begin{table}[t]
    \centering
    \scalebox{0.82}{
    \begin{tabular}{c|c|ccc}
        \toprule[1.0pt]
        \multirow{2}{*}{Models} & CLIP- & \multicolumn{3}{c}{CLIP-IQA~\cite{wang2022exploring}}  \\
        & Score$_\uparrow$~\cite{hessel2021clipscore} & Quality$_\uparrow$ & Colorful$_\uparrow$ & Sharp$_\uparrow$ \\
        \midrule
        RGBD2~\cite{lei2023rgbd2} & 0.2035 & 0.1279 & 0.2081 & 0.0126 \\
        \textbf{LucidDreamer} & \textbf{0.2110} & \textbf{0.6161} & \textbf{0.8453} & \textbf{0.5356} \\
        \bottomrule[1.0pt]
    \end{tabular}
    }
    \caption{\textbf{Quantitative comparison of generated scenes.}
    We quantitatively compare the results using CLIP-Score and CLIP-IQA with RGBD2.
    Our model shows better results on all metrics.
    }
    \label{tab:clip}
\end{table}

In \figref{fig:rgbd2-i1}, we observe that RGBD2 generates (ScanNet-style) images with similar styles regardless of the input image.
This remains consistent not only in the initial image but also throughout the following sequence as shown in \figref{fig:rgbd2-is}.
We believe the issue arises due to insufficient training data and domain limitations, highlighting the need for a model with sufficient generalization.
In contrast, our approach generates high-quality 3D scenes with careful consideration to harmonize well with the input RGB.
Moreover, LucidDreamer can generate scenes composed of high-resolution images while RGBD2 can only make $128\times128$-sized images, which is too small to use in real applications. 
We also document the quantitative results evaluated on CLIP-Score~\cite{hessel2021clipscore} and CLIP-IQA~\cite{wang2022exploring} in Table~\ref{tab:clip}.
We confirm that our model incorporates input conditions well, resulting in the creation of high-quality 3D scenes.

\begin{figure}[t]
    \centering
    \subfloat[Trained with mask]{\includegraphics[width=0.48\linewidth]{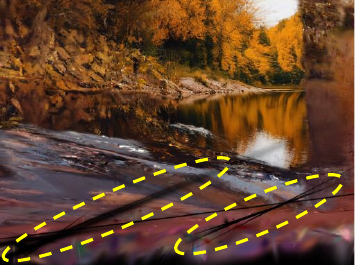}}
        \hfill
        \subfloat[Trained without mask]{\includegraphics[width=0.48\linewidth]{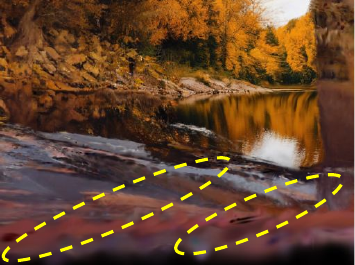}}
    \vspace{-2mm}
    \caption{
        \textbf{The effect of the mask during training.} 
        Training with valid masks helps prevent artifacts at the boundaries of the scene.
    }
    \label{fig:ab_mask}
    \vspace{-3mm}
\end{figure}

\paragraph{Ablations on design choices.}
We demonstrate ablation studies on the design choices of LucidDreamer. 
In table~\ref{tab:psnr}, we compare the effect of COLMAP initialization and the point cloud initialization provided by our method when learning Gaussian splatting. 
LucidDreamer achieved a high-quality scene within fewer iterations by offering a substantial amount of high-quality initialization points. 
\figref{fig:ab_mask} illustrates the effectiveness of masking the image while learning Gaussian splatting. 
It removes the black splinter from the black background, resulting in visually pleasing outcomes.

\begin{table}[t]
    \centering
    \scalebox{0.93}{
    \begin{tabular}{c|c|ccc}
        \toprule[1.0pt]
        \multirow{2}{*}{Iters} & Source of & \multicolumn{3}{c}{Metrics} \\
        & SfM points & PSNR$_\uparrow$ & SSIM$_\uparrow$ & LPIPS$_\downarrow$ \\
        \midrule
        \multirow{2}{*}{1000} & COLMAP & 23.15 & 0.7246 & 0.2910 \\
        & \textbf{LucidDreamer} & \textbf{32.59} & \textbf{0.9672} & \textbf{0.0272} \\
        \midrule
        \multirow{2}{*}{3000} & COLMAP & 30.87 & 0.9478 & 0.0353 \\
        & \textbf{LucidDreamer} & \textbf{33.80} & \textbf{0.9754} & \textbf{0.0178} \\
        \midrule
        \multirow{2}{*}{7000} & COLMAP & 32.52 & 0.9687 & 0.0208 \\
        & \textbf{LucidDreamer} & \textbf{34.24} & \textbf{0.9781} & \textbf{0.0164} \\
        \bottomrule[1.0pt]
    \end{tabular}
    }
    \caption{\textbf{Reconstruction quality according to the source of initial SfM points.}
    We use the initial point cloud generated by COLMAP~\cite{schoenberger2016sfm,schoenberger2016mvs} and compare the reconstruction results.
    Our model consistently shows better reconstruction metrics.
    }
    \label{tab:psnr}
\end{table}

\section{Conclusion}
In this work, we propose LucidDreamer, a novel pipeline for domain-free 3D scene generation.
By fully exploiting the power of large diffusion models, LucidDreamer is capable of generating high-quality scenes without the restriction of the target scene domain.
We first generate the point cloud from the input image and repeat `Dreaming' and `Alignment' algorithms to generate the multi-view consistent high-qulaity image and harmoniously integrate them to the existing point cloud in the 3D space.
After the construction is finished, the point cloud is converted to 3D Gaussian splats to enhance the quality of the 3D scene.
Extensive experiments show that LucidDreamer can consistently generate high-quality and diverse 3D scenes in various situations.

\end{document}